\documentclass[sigconf]{acmart}
\AtBeginDocument{%
  }

\usepackage{multirow}
\usepackage{pifont}
\usepackage[table]{xcolor}
\usepackage{subcaption}

\newcommand{\para}[1]{\smallskip\noindent\textit{#1.}\ }

\copyrightyear{2026}
\acmYear{2026}
\setcopyright{cc}
\setcctype{by}
\acmConference[CIKM '26]{Proceedings of the 35th ACM International Conference on Information and Knowledge Management}{November 07--11, 2026}{Rome, Italy}
\acmBooktitle{Proceedings of the 35th ACM International Conference on Information and Knowledge Management (CIKM '26), November 07--11, 2026, Rome, Italy}
\acmDOI{10.1145/3799682.3840662}
\acmISBN{979-8-4007-2539-5/2026/11}

\begin{document}

\title{Less is More: Concise Retrieval and Critic-based Re-ranking for Composed Image Retrieval}


\author{Yuqian Zheng}
\authornote{Equally contributed first authors}
\orcid{0009-0009-1683-1806}
\affiliation{%
\department{Department of Applied Artificial Intelligence}
  \institution{Sungkyunkwan University}
  \city{Seoul}
  \country{Republic of Korea}
}\email{zhengyuqian@g.skku.edu}

\author{Hyoju Ahn}
\authornotemark[1]
\orcid{0009-0006-6717-6889}
\affiliation{%
\department{Department of Immersive Media Engineering}
  \institution{Sungkyunkwan University}
  \city{Seoul}
  \country{Republic of Korea}
}\email{hyoxdo0717@skku.edu}

\author{Hyeongjin Ahn}
\authornote{Corresponding authors}
\orcid{0000-0001-8545-7867}
\affiliation{%
\department{Department of Applied Artificial Intelligence}
  \institution{Sungkyunkwan University}
  \city{Seoul}
  \country{Republic of Korea}
}\email{themilkywaydream@gmail.com}

\author{Eunil Park}
\authornotemark[2]
\correspondingauthor
\orcid{0000-0002-3177-3538}
\affiliation{%
\department{Department of Applied Artificial Intelligence}
  \institution{Sungkyunkwan University}
  \city{Seoul}
  \country{Republic of Korea}
}
\affiliation{%
	\institution{Jaume I University}
	\city{Castellon}
	\country{Spain}
}\email{eunilpark@skku.edu}

\renewcommand{\shortauthors}{Yuqian Zheng, Hyoju Ahn, Hyeongjin Ahn, and Eunil Park}

\begin{abstract}
Zero-shot composed image retrieval (ZS-CIR) aims to retrieve a target image based on a reference image and modification text. Recent approaches commonly rely on large vision-language models to generate target image captions for retrieval. However, these captions may include redundant semantic details, introducing noise when retrieving the target image and leading to suboptimal retrieval results. To address this issue, we propose \textit{Less-is-More CIR}, a two-stage training-free framework that combines Concise Retrieval and Critic Re-ranking. In the first stage, concise target-oriented captions are generated that preserve only the most discriminative modification semantics, thereby reducing semantic noise in the retrieval. In the second stage, a critic re-rank module further evaluates the retrieved candidates from the perspectives of modification consistency and reference preservation. Extensive experiments demonstrate the effectiveness and robustness of the proposed framework for the ZS-CIR. Our code is available at: \url{https://github.com/dxlabskku/LM-CIR}.
\end{abstract}



\maketitle

\section{Introduction}
Composed Image Retrieval (CIR)~\cite{Wen_2023, 10.1145/3474085.3475659} has recently attracted increasing attention owing to its wide applications in Internet search and e-commerce~\cite{saito2023pic2word, Chen_2020_CVPR}. The task aims to retrieve a target image from a database given a reference image and modification text that describes the desired changes. Because the query consists of both visual and textual information, the model must effectively understand and integrate these two modalities to accurately capture the user’s intent.

Prior work~\cite{Baldrati_2022_CVPR, Liu_2021_ICCV,10.5555/3692070.3694524} explored supervised approaches that rely on triplet annotations consisting of a reference image, modification text, and target image. However, such methods heavily depend on large-scale human-annotated data, which are costly to obtain. Moreover, distribution shifts across datasets often require training separate models for each dataset, leading to significant time and computational overheads. To alleviate these limitations, Zero-Shot Composed Image Retrieval (ZS-CIR) has been proposed to reduce the reliance on task-specific annotated triplets, making CIR more scalable and practical.

Recent ZS-CIR methods tend to avoid supervised training by leveraging large language models (LLMs) or large vision-language models (LVLMs) to better interpret the modification text and generate target image captions ~\cite{COTMR,osrcir,IP-CIR,CIReVL,SEIZE}. This caption is then encoded into the joint embedding space of CLIP~\cite{pmlr-v139-radford21a}, and retrieval is performed by ranking candidate images in the database according to the cosine similarity between their image embeddings and generated caption embedding. Depending on the caption generation process, existing methods can be broadly categorized into two categories. One-step methods~\cite{COTMR,osrcir} directly feed the reference image and modification text into an LVLM and generate the target image caption using chain-of-thought (CoT) prompting. In contrast, two-step methods~\cite{IP-CIR,CIReVL,SEIZE} first employ a pretrained image captioner to describe the reference image and then use an LLM to edit this description according to the modification text. In both cases, the quality of the generated target image captions plays a crucial role in determining the retrieval performance.

\begin{figure}[t]
    \centering
    \includegraphics[width=\columnwidth]{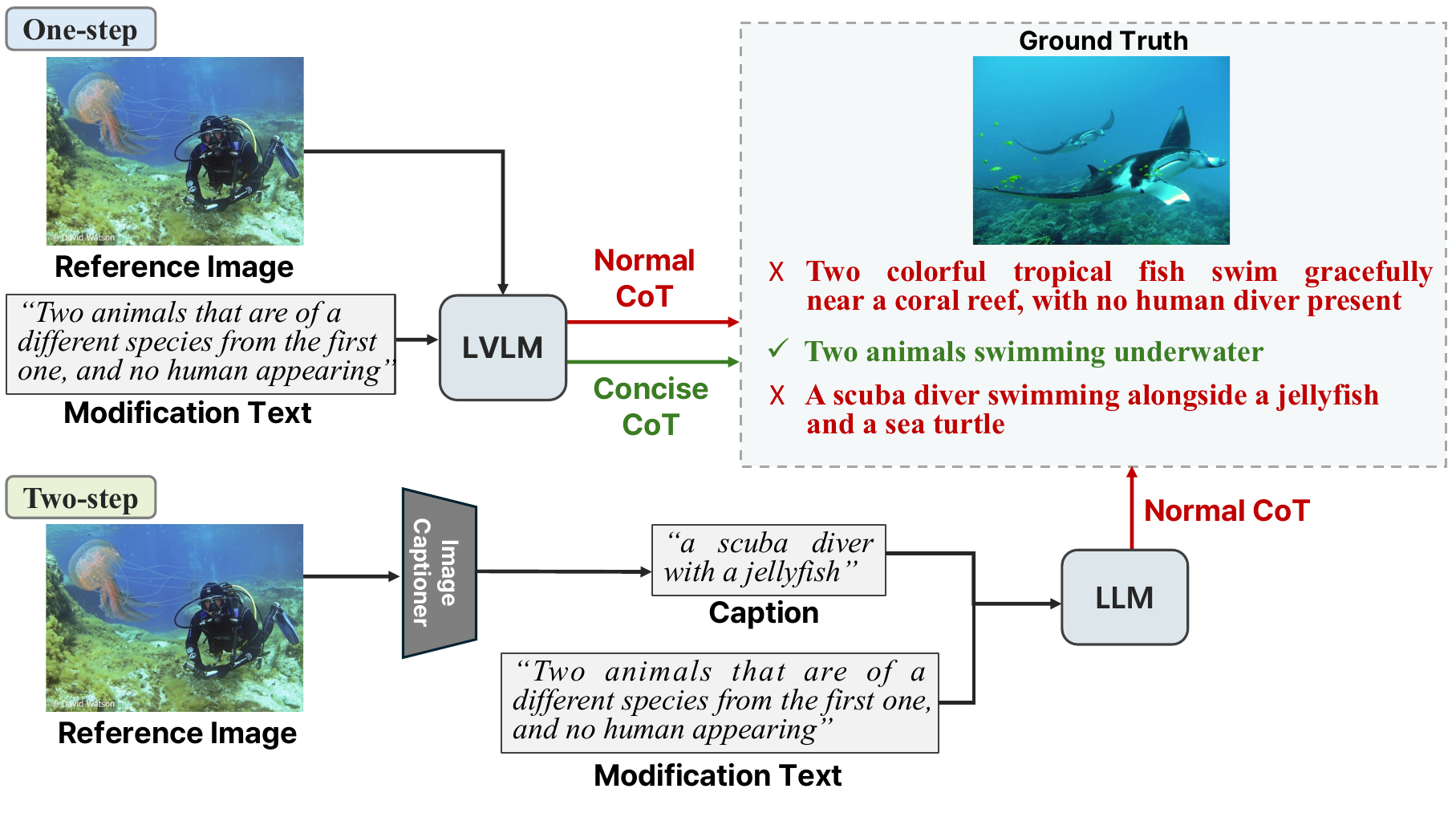}
    \caption{Comparison of target image captions generated by existing ZS-CIR methods and our concise reasoning strategy. Normal CoT introduces semantic noise, whereas concise reasoning generates cleaner target descriptions.}
    \Description{Comparison of target image captions generated by existing ZS-CIR methods and our concise reasoning strategy. Normal CoT introduces semantic noise, whereas concise reasoning generates cleaner target descriptions.}
    \label{fig:figure1}
\end{figure}

However, a common issue with existing approaches is that the generated captions tend to be overly verbose, frequently introducing irrelevant or hallucinated details that are not present in the target image. 
This excessive information introduces semantic noise and degrades the retrieval performance. 
Both the one- and two-step methods exhibit this issue.
As illustrated in Figure~\ref{fig:figure1}, generating target captions for CIR becomes particularly challenging when the modification text specifies abstract changes without concrete visual details. For example, because the modification text specifies adding “a different species” without indicating a concrete category, the generated captions hallucinate specific entities such as sea turtles or coral reefs. In addition, when the modification text requires that no human be present, the generated captions still describe a person or explicitly include negation expressions such as “no human.” 
Such descriptions introduce irrelevant or misleading signals that can adversely affect retrieval based on semantic similarity.

To address this issue, we propose a novel retrieval strategy, termed Concise Retrieval. This method explicitly constrains the LVLM via CoT prompting to generate concise target image captions that avoid unnecessary details unrelated to the target image. As shown in Figure~\ref{fig:figure1}, the resulting concise captions preserve only the core subject information from the reference image while appropriately incorporating the semantics of the modified text. In addition, when the modified text involves removal operations, our method avoids explicitly mentioning objects that should not appear in the target image, thereby reducing the noise introduced by negation. 
By suppressing redundant or hallucinated semantics, concise retrieval can produce more discriminative retrieval query captions and reduce mismatched semantic signals during text-image similarity computation.
Our experimental results demonstrate that when retrieval is performed solely based on text-image similarity, concise captions significantly outperform captions generated by existing methods. Considering that CIR is inherently a multimodal retrieval task, relying solely on text-based similarity may result in underutilization of visual information. Therefore, we further incorporate the reference image by computing the image-image similarity to complement the retrieval process. The final results of Concise Retrieval are obtained by combining text-image and image-image similarities.

However, overly concise captions may omit fine-grained attributes that are crucial for distinguishing between semantically similar candidates. To compensate for this limitation, we introduce a Critic Re-ranking mechanism that leverages an LVLM as a critic to perform fine-grained re-ranking of retrieved candidates. Specifically, the critic evaluates each candidate from two complementary perspectives: (1) EDIT, which measures whether the candidate correctly reflects the intended modification relative to the reference image; and (2) PRESERVE, which assesses whether the candidate retains key information from the reference image. The final ranking is obtained by combining the EDIT and PRESERVE scores. Based on this two-stage design, we propose a training-free ZS-CIR framework termed \textit{Less-is-More CIR} (LM-CIR). Our framework consistently outperforms existing methods across multiple benchmark datasets, achieving strong retrieval performance. The main contributions of this study are summarized as follows:

\begin{itemize}
    \item We propose Concise Retrieval, demonstrating the effectiveness of concise target image captions in ZS-CIR;

    \item We introduce Critic Re-ranking, which complements Concise Retrieval by enabling fine-grained ranking through EDIT and PRESERVE evaluation;

    \item We develop a training-free LM-CIR framework with strong performance on three CIR benchmarks.
\end{itemize}

\section{Related Work}
\subsection{Composed Image Retrieval}
CIR is a task that aims to retrieve a target image using a composed query consisting of visual and textual information~\cite{vo2019composing}. While early studies mainly developed supervised learning-based approaches~\cite{jang2024visual,jawade2025scot,Liu_2021_ICCV,10.5555/3692070.3694524}, their reliance on costly annotations and limited domain scalability have led to increasing attention on ZS-CIR, which leverages pre-trained models without task-specific triplet annotations.

Early ZS-CIR studies primarily focused on inversion-based approaches that map the reference image into the CLIP text embedding space~\cite{baldrati2023zero,lin2024fine,li2026modality,saito2023pic2word,chong2025leveraging}. For example, Pic2Word~\cite{saito2023pic2word} and SEARLE~\cite{baldrati2023zero} represent the reference image as a pseudo-word or textual token using image-caption pairs and combine it with the modification text to construct a composed query. Although these approaches avoid task-specific triplet annotations, they still require additional training on image-caption or unlabeled image datasets~\cite{baldrati2023zero,chong2025leveraging}. In addition, because the reference image is compressed into a limited number of pseudo tokens, their representation capacity is inherently constrained, making it challenging to explicitly capture the fine-grained visual details and complex visual attributes of the reference image~\cite{suo2024knowledge,jang2024spherical,lin2024fine}.

Accordingly, recent studies have actively explored training-free ZS-CIR methods that generate retrieval query captions by leveraging the reasoning capabilities of LLMs or LVLMs without additional training~\cite{COTMR,osrcir,IP-CIR,CIReVL,SEIZE,AutoCIR}. CIReVL~\cite{CIReVL} combines the reference image and modification text in the language space by captioning the reference image and recomposing it according to the modification text, and then uses the resulting target caption for CLIP-based retrieval.
In addition, OSrCIR~\cite{osrcir} generates a target description that jointly considers the reference image and modification intent through reflective CoT reasoning with an LVLM, while IP-CIR~\cite{IP-CIR} aims to mitigate the limitations of text-only retrieval by leveraging an imagined proxy of the target image. Overall, recent training-free ZS-CIR studies have evolved to generate captions that explicitly describe the target image using LLMs or LVLMs.

Nevertheless, relying heavily on generated captions introduces new challenges. Richly generated captions in existing caption-based, training-free ZS-CIR methods may include irrelevant attributes or hallucinated objects, introducing semantic noise into the CLIP embedding space. Moreover, most existing caption-based ZS-CIR studies release only inference code without providing the generated captions, thereby making exact reproduction difficult due to the inherent stochasticity of LVLM generation.


\subsection{LVLM-based Re-ranking and Critic}

In retrieval systems, re-ranking is widely adopted as a strategy that re-evaluates the top-ranked candidates retrieved in the initial retrieval stage using more expressive models~\cite{pandit2025evolution,zhou2025large,sasazawa2023text}. 
In CIR, initial retrieval typically relies on similarity search in a shared embedding space, enabling efficient search over large-scale candidate sets~\cite{weller2025theoretical,killingback2026scaling}. 
However, this similarity-based paradigm has limited ability to verify whether retrieved candidates satisfy fine-grained compositional constraints. This motivates re-ranking as a complementary stage for more detailed candidate assessment~\cite{liu2023candidate}.

LVLMs are well suited for re-ranking because they have recently shown strong capability in multimodal evaluation~\cite{li2024llms,zhou2025perception,chen2024mllm,Sun_2026_CVPR}.
By jointly understanding visual content and textual instructions, they are particularly suitable for vision-language evaluation tasks that require assessing visual consistency and instruction-conditioned relevance~\cite{xu2024lvlm,li2025survey,zhou2025perception}.

Existing ZS-CIR studies mainly use LVLMs for target caption generation, while only a few works explore candidate-level verification. For example, AutoCIR~\cite{AutoCIR} employs LVLM-based multi-agent collaboration to iteratively refine target captions and retrieval results through self-correction mechanisms. ImageScope~\cite{luo2025imagescope} leverages LVLMs to perform multistage retrieval refinement through semantic synthesis, predicate-level verification, and pairwise evaluation. However, these methods do not explicitly evaluate each retrieved candidate based on CIR-specific criteria. In contrast, our work focuses on critic-based re-ranking tailored to ZS-CIR, where modification consistency and reference preservation are used as explicit evaluation criteria to complement the semantic limitations of embedding-based retrieval.

\section{Method}
\subsection{Architecture Overview}
\label{sec:architecture}
\begin{figure*}[t]
    \centering

    \begin{subfigure}{0.95\textwidth}
        \centering
        \includegraphics[width=\textwidth]{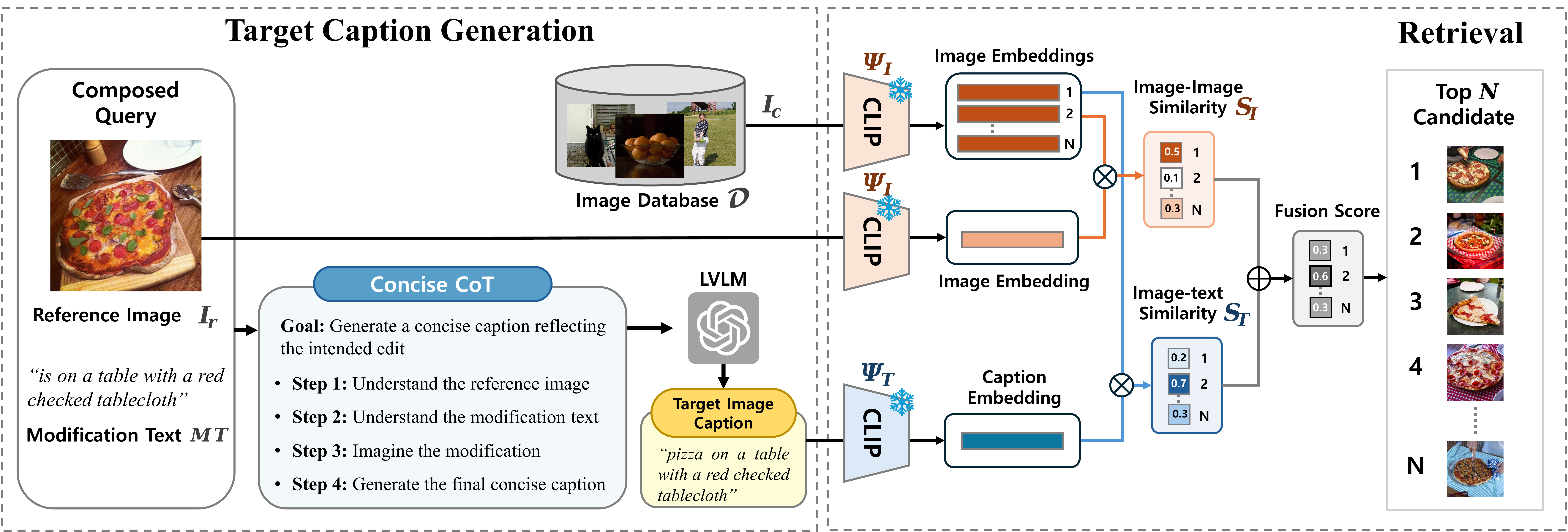}
        \caption{Stage 1. Concise Retrieval}
    \end{subfigure}

    \vspace{0.1em}

    \begin{subfigure}{0.95\textwidth}
        \centering
        \includegraphics[width=\textwidth]{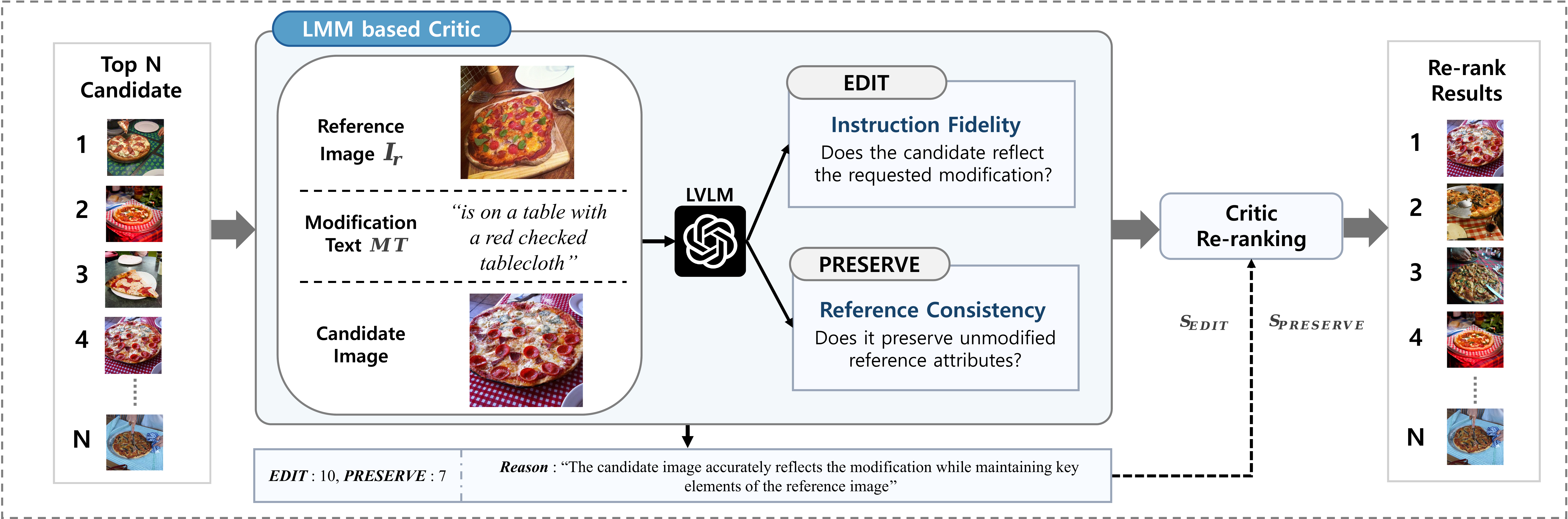}
        \caption{Stage 2. Critic Re-ranking}
    \end{subfigure}

    \caption{Overview of the two-stage LM-CIR framework. (a) In Stage 1, an LVLM generates a concise target image caption and retrieves $N$ candidates using fused image-image and text-image similarity scores. (b) In Stage 2, the Top-20 candidates are re-ranked based on EDIT and PRESERVE scores predicted by the LVLM critic.}
    \Description{Overview of the two-stage LM-CIR framework. (a) In Stage 1, an LVLM generates a concise target image caption and retrieves $N$ candidates using fused image-image and text-image similarity scores. (b) In Stage 2, the Top-20 candidates are re-ranked based on EDIT and PRESERVE scores predicted by the LVLM critic.}
    \label{fig:framework}
\end{figure*}

As illustrated in Figure~\ref{fig:framework}, the proposed LM-CIR framework consists of two stages: \textit{Concise Retrieval} and \textit{Critic Re-Ranking}. In the first stage, for each query, an LVLM is employed to generate a concise target image caption conditioned on the reference image and modification text. The generated caption is subsequently encoded into the shared vision-language embedding space of CLIP, where the cosine similarity between the caption embedding and the embeddings of candidate images is computed as the text-image similarity score. To further exploit the visual information contained in the reference image, image-image similarity between the reference image and candidate images is additionally incorporated. The two similarity measures are fused to produce an initial retrieval ranking.
In the second stage, we utilize the LVLM as a critic to re-rank the Top-$K$ candidates ($K=20$) retrieved from the first stage. Each candidate image is evaluated from two perspectives: whether it satisfies the intended modification (\textbf{EDIT}) and whether it preserves the key information of the reference image (\textbf{PRESERVE}). The final ranking results are then obtained based on the combined scores.

\subsection{Stage 1: Concise Retrieval}
As described in Section~\ref{sec:architecture}, the first stage aims to obtain high-quality preliminary retrieval results by generating concise and semantically accurate target image captions and combining them with multimodal similarity computations. Given a reference image $I_r$ and a modification text $MT$, both inputs are jointly fed into an LVLM to generate a concise caption, referred to as the $Concise\_Caption$, under a specifically designed Concise CoT prompting strategy:
\begin{equation}
Concise\_Caption = \mathrm{LVLM}(I_r, MT \mid \text{Concise CoT})
\end{equation}

The Concise CoT prompt guides the model through a structured reasoning process, as illustrated in Figure~\ref{fig:Prompt1}. Specifically, the procedure consists of four steps: (1) identifying the primary objects and their relationships in the reference image; (2) parsing the modification text to determine the edited objects and attributes; (3) semantically inferring the appearance of the target image according to the modification instructions; and (4) generating the final concise caption.

Unlike existing approaches that often produce verbose and detail-rich captions, our method explicitly introduces conciseness constraints to suppress irrelevant semantic information that may hinder retrieval. Specifically, the constraints are defined according to the following principles:
\begin{itemize}
    \item \textbf{Avoiding negative expressions}: for negation or removal operations (e.g., ``not green'' or ``no people''), the corresponding attributes or objects are omitted rather than explicitly described in the generated caption;
    \item \textbf{Avoiding semantic hallucination}: when the modification text does not explicitly specify categories or attributes (e.g., ``an animal''), the caption remains semantically generic instead of introducing uncertain or overly specific details;  
    \item \textbf{Maintaining minimally sufficient descriptions}: only the core subjects and attributes relevant to the manipulation are retained, while redundant details and unrelated information are excluded.
\end{itemize}

To further improve the generation stability and reinforce the desired behavior of concise caption generation, five in-context examples are additionally incorporated into the prompt, as illustrated in Figures~\ref{fig:prompt_examples1} and~\ref{fig:prompt_examples2}.
After obtaining the $Concise\_Caption$, it is projected into the shared embedding space of CLIP, where the text-image similarity is computed with each candidate image $I_c \in \mathcal{D}$, with $\mathcal{D}$ denoting the candidate image database:
\begin{equation}
S_{TI}(I_c) = \frac{\Psi_I(I_c)^\top \Psi_T(\text{Concise\_Caption})}
{\|\Psi_I(I_c)\| \cdot \|\Psi_T(\text{Concise\_Caption})\|}
\end{equation}
where $\Psi_I(\cdot)$ and $\Psi_T(\cdot)$ denote the CLIP image and text encoders, respectively. To further leverage the visual semantics of the reference image, the image-image similarity between the reference image $I_r$ and each candidate image is computed as:
\begin{equation}
S_{II}(I_c) = \frac{\Psi_I(I_c)^\top \Psi_I(I_r)}
{\|\Psi_I(I_c)\| \cdot \|\Psi_I(I_r)\|}
\end{equation}

The two similarity scores are subsequently fused through weighted summation to obtain the overall retrieval score for each candidate image as follows:
\begin{equation}
S(I_c) = S_{TI}(I_c) + \alpha \cdot S_{II}(I_c)
\end{equation}
where $\alpha$ denotes the weighting coefficient that balances the two similarity terms. Candidate images are ranked according to the combined score to produce preliminary retrieval results, which are then passed to the subsequent Critic Re-ranking stage.

\subsection{Stage 2: Critic Re-ranking}
Although concise captions reduce semantic noise in the initial retrieval stage, they may omit fine-grained visual attributes that are useful for distinguishing between visually similar candidates. Moreover, embedding-based retrieval does not explicitly verify whether each retrieved candidate satisfies the composed query. In CIR, a valid candidate must meet two complementary requirements: it should reflect the intended modification while preserving the key information of the reference image. 
To address this, we introduce an LVLM-based critic, guided by the prompt shown in Fig.~\ref{fig:Prompt2}, that directly evaluates the retrieved candidates before producing the final ranking.
Owing to its strong multimodal understanding and reasoning capabilities, the LVLM critic can jointly analyze visual details and textual modification instructions, making it well-suited for fine-grained evaluation in CIR.

Given a reference image $I_r$, modification text $MT$, and candidate image $I_c$ retrieved from the first-stage retriever, the critic estimates the suitability of $I_c$ from two perspectives. The first is instruction fidelity, which measures whether the candidate correctly reflects the modification described in $MT$ with respect to $I_r$. The second is reference consistency, which measures whether the candidate preserves the visual attributes of $I_r$ that are not intended to be changed. These two criteria correspond to the EDIT and PRESERVE scores, respectively.

For each candidate, the LVLM critic receives $(I_r, MT, I_c)$ as input and outputs two scores, $S_{EDIT}, S_{PRESERVE} \in {0,1,\ldots,10}$, along with brief explanations of the assigned scores. The $S_{EDIT}$ evaluates how well the requested modification is applied, while the $S_{PRESERVE}$ evaluates how well the unchanged attributes of the reference image are maintained. By separating these two aspects, a critic can identify candidates that remain visually similar to the reference image but fail to apply the requested modification, as well as candidates that reflect the modification but lose important reference information.

We compute the final critic score for each candidate as:
\begin{equation}
S_{CRITIC} = \lambda \cdot S_{EDIT} + (1-\lambda) \cdot S_{PRESERVE}
\end{equation}
where $\lambda$ controls the trade-off between modification fidelity and reference preservation.
Because fine-grained semantic verification is computationally expensive, we apply the critic-based re-ranking process only to the top-$N$ candidates from the initial ranking produced by the first-stage retriever. These candidates are re-ranked in descending order according to $S_{CRITIC}$, whereas the remaining candidates retain their original retrieval order.

\section{Experiments}

\subsection{Baselines and Datasets}
To validate the effectiveness of the proposed approach, we compare it with several recent representative ZS-CIR methods, including \textit{CIReVL}~\cite{CIReVL}, \textit{OSrCIR}~\cite{osrcir}, \textit{AutoCIR}~\cite{AutoCIR}, \textit{ImageScope}~\cite{luo2025imagescope}, \textit{SEIZE}~\cite{SEIZE}, \textit{CoTMR}~\cite{COTMR}, \textit{IP-CIR}~\cite{IP-CIR}, and \textit{G-MIXER}~\cite{G-MIXER}. All selected baselines are training-free ZS-CIR methods. Among them, \textit{CIReVL}, \textit{OSrCIR}, and \textit{AutoCIR} primarily rely on target image descriptions to perform semantic matching with candidate images during retrieval. In contrast, \textit{ImageScope}, \textit{SEIZE}, \textit{CoTMR}, \textit{IP-CIR}, and \textit{G-MIXER} leverage richer retrieval cues beyond target image captions, such as reference images and include/exclude information extracted by LVLMs, thereby enhancing the retrieval performance in complex compositional scenarios.

We evaluate the proposed method on three widely used ZS-CIR benchmarks: FashionIQ~\cite{Wu_2021_CVPR}, CIRR~\cite{Liu_2021_ICCV}, and CIRCO~\cite{baldrati2023zero}. These benchmarks represent different retrieval settings and pose distinct challenges for the CIR models. Among them, CIRCO is the first CIR benchmark that provides multiple ground-truth target images for each query, enabling a more comprehensive evaluation of the retrieval performance in open-world settings. The CIRR is a single-target retrieval benchmark that introduces a subset setting in which retrieval is performed within a restricted candidate pool. 
FashionIQ is a single-target retrieval benchmark specifically designed for fine-grained fashion retrieval. It contains three apparel categories, namely \textit{Dress}, \textit{Shirt}, and \textit{Toptee}, and primarily evaluates the model’s ability to understand attribute modifications and compositional semantics in fashion-image data.

For the evaluation metrics, we follow the standard protocols commonly adopted in prior ZS-CIR studies. For CIRCO, which contains multiple target images per query, we report the mean Average Precision (mAP@$k$). For CIRR and FashionIQ, Recall@$k$ (R@$k$) is adopted as the primary evaluation metric. In addition, Recall$_{sub}$@$k$ is used under the subset setting of CIRR to measure the retrieval performance within the restricted candidate set.
It should be noted that the ground-truth annotations of the CIRCO and CIRR test sets are unavailable publicly. Instead, both benchmarks provide official online evaluation servers for performance assessment. 
In contrast, because the official evaluation server of FashionIQ is no longer available, we follow many previous studies and conduct a local evaluation on its validation set.

\subsection{Implementation Details}
The entire framework is implemented using PyTorch~\cite{10.5555/3454287.3455008} and conducted on a single NVIDIA H200 GPU. For the retrieval model, we adopt the ViT-G/14, ViT-B/32, and ViT-L/14 variants from OpenCLIP~\cite{ilharco_gabriel_2021_5143773}. In Stage 1, we use GPT-5.1~\cite{openai_gpt51_2025} as the LVLM through API-based inference. In the concise CoT prompting strategy, the rules for generating concise captions are generally shared across different datasets, while the in-context examples are designed according to the characteristics of each dataset. Specifically, for CIRR and CIRCO, we employ a set of generic examples, whereas for FashionIQ, the examples are specifically tailored to the fashion domain. In addition, the fusion weight $\alpha$ for combining text-image similarity and image-image similarity is set to 0.2 on CIRR and CIRCO, and 0.3 on FashionIQ, where the values are determined based on validation performance.

In Stage 2, we employ Qwen2.5-VL-32B-Instruct as the LVLM critic. For each query, the Top-20 candidates retrieved in Stage 1 are evaluated using the reference image, modification text, and candidate image as inputs. The critic assigns \textbf{EDIT} and \textbf{PRESERVE} scores to each candidate, which are interpolated with \(\lambda=0.7\) to obtain the final critic score. The interpolation weight \(\lambda\) is empirically selected based on preliminary experiments on the validation dataset. To ensure reproducibility, we use deterministic decoding with a temperature set to 0.

\subsection{Quantitative Results}
\para{CIRCO}
Our proposed framework consistently achieves state-of-the-art performance across all CLIP backbone models, as shown in Table~\ref{tab:main_results}. In particular, the results after the critic re-rank stage significantly outperform existing methods across all evaluation metrics. Using ViT-G/14 as an example, \textit{Ours - Critic} achieves 42.79\% in mAP@5 and 43.85\% in mAP@50, surpassing the previous best method IP-CIR by +10.04\% and +5.82\%, respectively. Notably, the larger improvement observed at mAP@5 suggests that our framework is especially effective at ranking the most relevant target images within the top retrieval positions.
This improvement is likely because the critic re-rank stage can better distinguish highly similar candidate images through fine-grained compositional reasoning beyond coarse embedding-level matching.

\begin{table*}[t]
\centering
\setlength{\tabcolsep}{6pt}
\renewcommand{\arraystretch}{0.7}
\caption{Performance comparison on the CIRCO and CIRR test benchmarks. $sub$ denotes the CIRR subset retrieval task. Moreover, the best and second-best results are indicated by \textbf{bold} and \underline{underlined} text, respectively.}
\label{tab:main_results}

\begin{tabular}{lc|cccc|cccc|ccc}
\toprule
\multicolumn{2}{c|}{\textbf{Benchmark}}
& \multicolumn{4}{c|}{\textbf{CIRCO}}
& \multicolumn{7}{c}{\textbf{CIRR}} \\
\midrule
\multicolumn{2}{c|}{\textbf{Metric}}
& \multicolumn{4}{c|}{\textbf{mAP@$k$}}
& \multicolumn{4}{c|}{\textbf{Recall@$k$}}
& \multicolumn{3}{c}{\textbf{Recall$_{sub}$@$k$}} \\
\midrule
\textbf{Backbone} & \textbf{Method}
& \textbf{$k$=5} & \textbf{$k$=10} & \textbf{$k$=25} & \textbf{$k$=50}
& \textbf{$k$=1} & \textbf{$k$=5} & \textbf{$k$=10} & \textbf{$k$=50}
& \textbf{$k$=1} & \textbf{$k$=2} & \textbf{$k$=3} \\
\midrule

\multirow{8}{*}{ViT-G/14}
& CIReVL~\cite{CIReVL} & 26.77 & 27.59 & 29.96 & 31.03 & 34.65 & 64.29 & 75.06 & 91.66 & 67.95 & 84.87 & 93.21 \\
& OSrCIR~\cite{osrcir} & 30.47 & 31.14 & 35.03 & 36.59 & 37.26 & 67.25 & 77.33 & -- & 69.22 & 85.28 & 93.55 \\
& SEIZE~\cite{SEIZE} & 32.46 & 33.77 & 36.46 & 37.55 & 38.87 & 69.42 & 79.42 & --    & 74.15 & 89.23 & 95.71 \\
& CoTMR~\cite{COTMR} & 32.23 & 32.72 & 35.60 & 36.83 & 36.36 & 67.52 & 77.82 & 93.99 & 71.19 & 86.34 & 93.87 \\
& IP-CIR~\cite{IP-CIR} & \underline{32.75} & \underline{34.26} & \underline{36.86} & \underline{38.03} & 39.25 & 70.07 & 80.00 & \cellcolor{pink!25}\textbf{94.89} & 69.95 & 86.87 & 94.22 \\
& G-MIXER~\cite{G-MIXER} & 31.79 & 32.54 & 35.49 & 36.87 & 39.18 & 69.83 & 79.35 & --    & 72.36 & 87.25 & 93.85 \\
\cmidrule(lr){2-13}
& \textit{Ours - Concise}  
& 31.93 & 32.97 & 35.64 & 36.67 
& \underline{41.33} & \underline{70.80} & \underline{81.33} & 94.39
& \underline{76.36} & \underline{90.63} & \underline{96.15} \\
& \textit{Ours - Critic}   
& \cellcolor{pink!25}\textbf{42.79} 
& \cellcolor{pink!25}\textbf{43.04} 
& \cellcolor{pink!25}\textbf{43.85} 
& \cellcolor{pink!25}\textbf{43.85} 
& \cellcolor{pink!25}\textbf{47.47} 
& \cellcolor{pink!25}\textbf{77.04} 
& \cellcolor{pink!25}\textbf{83.81} 
& \underline{94.39} 
& \cellcolor{pink!25}\textbf{78.05} 
& \cellcolor{pink!25}\textbf{91.69} 
& \cellcolor{pink!25}\textbf{96.53} \\
\midrule

\multirow{10}{*}{ViT-L/14}
& CIReVL~\cite{CIReVL} & 18.57 & 19.01 & 20.89 & 21.80 & 24.55 & 52.31 & 64.92 & 86.34 & 59.54 & 79.88 & 89.69 \\
& OSrCIR~\cite{osrcir} & 23.87 & 25.33 & 27.84 & 28.97 & 29.45 & 57.68 & 69.86 & -- & 62.12 & 81.92 & 91.10 \\
& AutoCIR~\cite{AutoCIR} & 24.05 & 25.14 & 27.35 & 28.36 & 31.81 & 61.95 & 73.86 & 92.07 & 67.21 & 84.89 & 93.13 \\
& ImageScope~\cite{luo2025imagescope} & 28.36 & 29.23 & 30.81 & 31.88 & \underline{39.37} & 67.54 & 78.05 & \underline{92.94} & \underline{76.36} & \underline{89.40} & \underline{95.21} \\
& SEIZE~\cite{SEIZE} & 24.98 & 25.82 & 28.24 & 29.35 & 28.65 & 57.16 & 69.23 & --    & 66.22 & 84.05 & 92.34 \\
& CoTMR~\cite{COTMR} & 27.61 & 28.22 & 30.61 & 31.70 & 35.02 & 64.75 & 76.18 & 92.51 & 69.39 & 85.75 & 93.33 \\
& IP-CIR~\cite{IP-CIR} & 26.43 & 27.41 & 29.87 & 31.07 & 29.76 & 58.82 & 71.21 & 90.41 & 62.48 & 81.64 & 90.89 \\
& G-MIXER~\cite{G-MIXER} & 28.29 & 29.04 & 31.44 & 32.39 & 37.42 & 67.69 & 78.58 & --    & 71.88 & 87.04 & 92.82 \\
\cmidrule(lr){2-13}
& \textit{Ours - Concise}  
& \underline{29.01} & \underline{29.62} & \underline{32.05} & \underline{33.17} 
& 38.39 & \underline{69.23} & \underline{80.17} & 94.27 
& 70.24 & 87.18 & 94.12 \\
& \textit{Ours - Critic}   
& \cellcolor{pink!25}\textbf{39.82} 
& \cellcolor{pink!25}\textbf{39.61} 
& \cellcolor{pink!25}\textbf{40.74} 
& \cellcolor{pink!25}\textbf{41.86} 
& \cellcolor{pink!25}\textbf{47.93} 
& \cellcolor{pink!25}\textbf{77.37} 
& \cellcolor{pink!25}\textbf{84.72} 
& \cellcolor{pink!25}\textbf{94.27} 
& \cellcolor{pink!25}\textbf{77.76} 
& \cellcolor{pink!25}\textbf{91.33} 
& \cellcolor{pink!25}\textbf{96.55} \\
\midrule

\multirow{9}{*}{ViT-B/32}
& CIReVL~\cite{CIReVL} & 14.94 & 15.42 & 17.00 & 17.82 & 23.94 & 52.51 & 66.00 & 86.95 & 60.17 & 80.05 & 90.19 \\
& OSrCIR~\cite{osrcir} & 18.04 & 19.17 & 20.94 & 21.85 & 25.42 & 54.54 & 68.19 & -- & 62.31 & 80.86 & 91.13 \\
& AutoCIR~\cite{AutoCIR} & 18.82 & 19.41 & 21.38 & 22.32 & 30.53 & 59.42 & 72.19 & 91.47 & 65.11 & 84.02 & 92.70 \\
& ImageScope~\cite{luo2025imagescope} & \underline{25.26} & \underline{25.82} & \underline{27.15} & \underline{28.11} & \underline{38.43} & 66.27 & 76.96 & \underline{91.83} & \underline{75.93} & \underline{89.21} & \underline{94.63} \\
& SEIZE~\cite{SEIZE} & 19.04 & 19.64 & 21.55 & 22.49 & 27.47 & 57.42 & 70.17 & --    & 65.59 & 84.48 & 92.77 \\
& CoTMR~\cite{COTMR} & 22.23 & 22.78 & 24.68 & 25.74 & 31.50 & 60.80 & 73.04 & 91.06 & 66.61 & 84.50 & 92.55 \\
& G-MIXER~\cite{G-MIXER} & 21.33 & 22.00 & 24.00 & 24.98 & 35.18 & 66.19 & 76.82 & --    & 69.98 & 85.78 & 92.92 \\
\cmidrule(lr){2-13}
& \textit{Ours - Concise}  
& 24.00 & 24.30 & 26.30 & 27.20 
& 36.75 & \underline{68.19} & \underline{79.21} & 94.75 
& 68.51 & 86.02 & 93.83 \\
& \textit{Ours - Critic}   
& \cellcolor{pink!25}\textbf{36.53} 
& \cellcolor{pink!25}\textbf{35.69} 
& \cellcolor{pink!25}\textbf{36.48} 
& \cellcolor{pink!25}\textbf{37.38} 
& \cellcolor{pink!25}\textbf{46.92} 
& \cellcolor{pink!25}\textbf{75.45} 
& \cellcolor{pink!25}\textbf{83.16} 
& \cellcolor{pink!25}\textbf{94.75} 
& \cellcolor{pink!25}\textbf{77.47} 
& \cellcolor{pink!25}\textbf{91.49} 
& \cellcolor{pink!25}\textbf{96.43} \\

\bottomrule
\end{tabular}
\end{table*}


Moreover, even without the critic re-rank stage, the first-stage retrieval module, \textit{Ours - Concise}, already achieves highly competitive and, in several settings, superior performance compared with existing approaches. For instance, on CIRCO with ViT-L/14, \textit{Ours - Concise} achieves 33.17\% mAP@50, outperforming G-MIXER (32.39\%) and ImageScope (31.88\%). 
These results demonstrate the effectiveness of concise retrieval in reducing semantic noise.

\para{CIRR}
Our framework demonstrates strong and consistent performance across different CLIP backbones and for most evaluation metrics. \textit{Ours - Critic} achieves competitive results on both Recall@$k$ and Recallsub@$k$. For example, under ViT-G/14, \textit{Ours - Critic} achieves 47.47\% Recall@1, surpassing the previous best method IP-CIR by +8.22\%. 
In the subset retrieval setting, the improvement margin is comparatively smaller because the candidate pool is considerably smaller, thereby reducing retrieval ambiguity. Nevertheless, our method still achieves 78.05\% Recallsub@1, outperforming the previous best method SEIZE by +3.9\%. 


\para{FashionIQ}
As shown in Table~\ref{tab:fiq_results}, under ViT-L/14 and ViT-B/32, both \textit{Ours - Concise} and \textit{Ours - Critic} outperform the existing methods on nearly all metrics. Under ViT-G/14, although our framework achieves the best performance on the Dress subset, the results on Shirt and Tops\&Tee are slightly lower than the strongest baselines. We conjecture that FashionIQ relies heavily on fine-grained garment descriptions, while a large number of visually similar clothing images exist in the retrieval pool. In such cases, concise captions may omit subtle attribute details that are useful for distinguishing highly similar fashion items. 
Moreover, because the critic re-rank stage is applied only to the Top-20 retrieved candidates, it cannot recover the ground-truth target when it is initially ranked too low. 
Nevertheless, our framework still maintains a highly competitive overall performance across all subsets, and the advantage of concise caption retrieval becomes more evident when using relatively lightweight CLIP backbones.

\begin{table*}[t]
\centering
\setlength{\tabcolsep}{10pt}
\renewcommand{\arraystretch}{0.7}
\caption{Performance comparison on the FashionIQ validation benchmark. The best and second-best results are indicated by \textbf{bold} and \underline{underlined} text, respectively.}
\label{tab:fiq_results}
\scalebox{1}{
\begin{tabular}{lc|cc|cc|cc|cc}
\toprule
\multirow{2}{*}{\textbf{Backbone}} 
& \multirow{2}{*}{\textbf{Method}}
& \multicolumn{2}{c|}{\textbf{Shirt}}
& \multicolumn{2}{c|}{\textbf{Dress}}
& \multicolumn{2}{c|}{\textbf{Toptee}}
& \multicolumn{2}{c}{\textbf{Avg.}} \\
\cmidrule(lr){3-4}
\cmidrule(lr){5-6}
\cmidrule(lr){7-8}
\cmidrule(lr){9-10}
& 
& \textbf{R@10} & \textbf{R@50}
& \textbf{R@10} & \textbf{R@50}
& \textbf{R@10} & \textbf{R@50}
& \textbf{R@10} & \textbf{R@50} \\
\midrule

\multirow{9}{*}{ViT-G/14}
& CIReVL~\cite{CIReVL} & 33.71 & 51.42 & 27.07 & 49.53 & 35.80 & 56.14 & 32.19 & 52.36 \\
& OSrCIR~\cite{osrcir} & 38.65 & 54.71 & 33.02 & 54.78 & 41.04 & 61.83 & 37.57 & 57.11 \\
& AutoCIR~\cite{AutoCIR} & 36.36 & 55.84 & 26.18 & 47.69 & 37.28 & 60.38 & 33.27 & 54.63 \\
& SEIZE~\cite{SEIZE} & 43.60 & \underline{65.42} & 39.61 & 61.02 & 45.94 & \underline{71.12} & 43.05 & \underline{65.85} \\
& CoTMR~\cite{COTMR} & 38.32 & 62.24 & 34.51 & 57.36 & 41.90 & 64.30 & 38.25 & 61.32 \\
& IP-CIR~\cite{IP-CIR} 
& \cellcolor{pink!25}\textbf{48.04} 
& \cellcolor{pink!25}\textbf{66.68} 
& 39.02 
& \underline{61.03} 
& \cellcolor{pink!25}\textbf{50.18} 
& \cellcolor{pink!25}\textbf{71.14} 
& \underline{45.74} 
& \cellcolor{pink!25}\textbf{66.28} \\
& G-MIXER~\cite{G-MIXER} & 39.65 & 59.61 & 34.71 & 58.85 & 44.77 & 67.47 & 39.71 & 61.98 \\
\cmidrule(lr){2-10}
& \textit{Ours - Concise}  
& 44.65 
& 63.44 
& \underline{40.41} 
& 63.21 
& 46.97 
& 67.06 
& 44.01 
& 64.57 \\
& \textit{Ours - Critic}   
& \underline{47.15} 
& 63.44 
& \cellcolor{pink!25}\textbf{41.20} 
& \cellcolor{pink!25}\textbf{63.21} 
& \underline{48.90} 
& 67.06 
& \cellcolor{pink!25}\textbf{45.75} 
& 64.57 \\
\midrule

\multirow{9}{*}{ViT-L/14}
& CIReVL~\cite{CIReVL} & 29.49 & 47.40 & 24.79 & 44.76 & 31.36 & 53.65 & 28.55 & 48.57 \\
& OSrCIR~\cite{osrcir} & 33.17 & 52.03 & 29.70 & 51.81 & 36.92 & 59.27 & 33.26 & 54.37 \\
& AutoCIR~\cite{AutoCIR} & 34.00 & 53.43 & 24.94 & 45.81 & 33.10 & 55.58 & 30.68 & 51.60 \\
& ImageScope~\cite{luo2025imagescope} & --    & --    & --    & --    & --    & --    & 31.36 & 50.78 \\
& SEIZE~\cite{SEIZE} & 33.04 & 53.22 & 30.93 & 50.76 & 35.57 & 58.64 & 33.18 & 54.21 \\
& CoTMR~\cite{COTMR} & 35.43 & 54.91 & 31.18 & 55.04 & 38.55 & 61.33 & 35.05 & 57.09 \\
& G-MIXER~\cite{G-MIXER} & 40.87 & \underline{60.35} & 37.98 & \underline{60.93} & \underline{46.91} & \cellcolor{pink!25}\textbf{66.14} & 41.92 & \underline{62.47} \\
\cmidrule(lr){2-10}
& \textit{Ours - Concise}  
& \underline{42.10} 
& 60.79 
& \underline{39.27} 
& 61.73 
& 45.95 
& 65.58
& \underline{42.44} 
& 62.70 \\
& \textit{Ours - Critic}   
& \cellcolor{pink!25}\textbf{44.46} 
& \cellcolor{pink!25}\textbf{60.79} 
& \cellcolor{pink!25}\textbf{40.60} 
& \cellcolor{pink!25}\textbf{61.73} 
& \cellcolor{pink!25}\textbf{48.24} 
& \underline{65.58} 
& \cellcolor{pink!25}\textbf{44.43} 
& \cellcolor{pink!25}\textbf{62.70} \\
\midrule

\multirow{9}{*}{ViT-B/32}
& CIReVL~\cite{CIReVL} & 28.36 & 47.84 & 25.29 & 46.36 & 31.21 & 53.85 & 28.29 & 49.35 \\
& OSrCIR~\cite{osrcir} & 31.16 & 51.13 & 29.35 & 50.37 & 36.51 & 58.71 & 32.34 & 53.40 \\
& AutoCIR~\cite{AutoCIR} & 32.43 & 51.67 & 26.52 & 46.36 & 33.96 & 56.09 & 30.97 & 51.37 \\
& ImageScope~\cite{luo2025imagescope} & --    & --    & --    & --    & --    & --    & 31.42 & 50.80 \\
& SEIZE~\cite{SEIZE} & 29.38 & 47.97 & 25.37 & 46.84 & 32.07 & 54.78 & 28.94 & 49.86 \\
& CoTMR~\cite{COTMR} & 33.42 & 53.93 & 31.09 & 54.54 & 38.40 & 61.14 & 34.30 & 56.54 \\
& G-MIXER~\cite{G-MIXER} & 37.24 & \underline{55.99} & 36.39 & \underline{58.21} & \underline{45.23} & \underline{64.10} & 39.62 & \underline{59.43} \\
\cmidrule(lr){2-10}
& \textit{Ours - Concise}  
& \underline{37.44} 
& 58.44 
& \underline{36.84} 
& 60.88 
& 44.77 
& 65.02 
& \underline{39.68} 
& 61.45 \\
& \textit{Ours - Critic}   
& \cellcolor{pink!25}\textbf{39.79} 
& \cellcolor{pink!25}\textbf{58.44} 
& \cellcolor{pink!25}\textbf{38.40} 
& \cellcolor{pink!25}\textbf{60.88} 
& \cellcolor{pink!25}\textbf{46.46} 
& \cellcolor{pink!25}\textbf{65.02} 
& \cellcolor{pink!25}\textbf{41.55} 
& \cellcolor{pink!25}\textbf{61.45} \\
\bottomrule
\end{tabular}
}
\end{table*}

Across all three benchmarks, the consistent gains from \textit{Ours - Concise} to \textit{Ours - Critic} suggest that relying solely on similarity matching in CLIP’s joint vision-language embedding space of CLIP is insufficient. Although concise retrieval can construct high-quality candidate sets, it becomes less discriminative when many visually similar candidates exist in the retrieval pool. In such cases, the critic re-rank stage leverages the reasoning capability of the LVLM to better distinguish subtle compositional differences, leading to more accurate retrieval results.

\begin{figure}[t]
    \centering
    \includegraphics[width=\columnwidth]{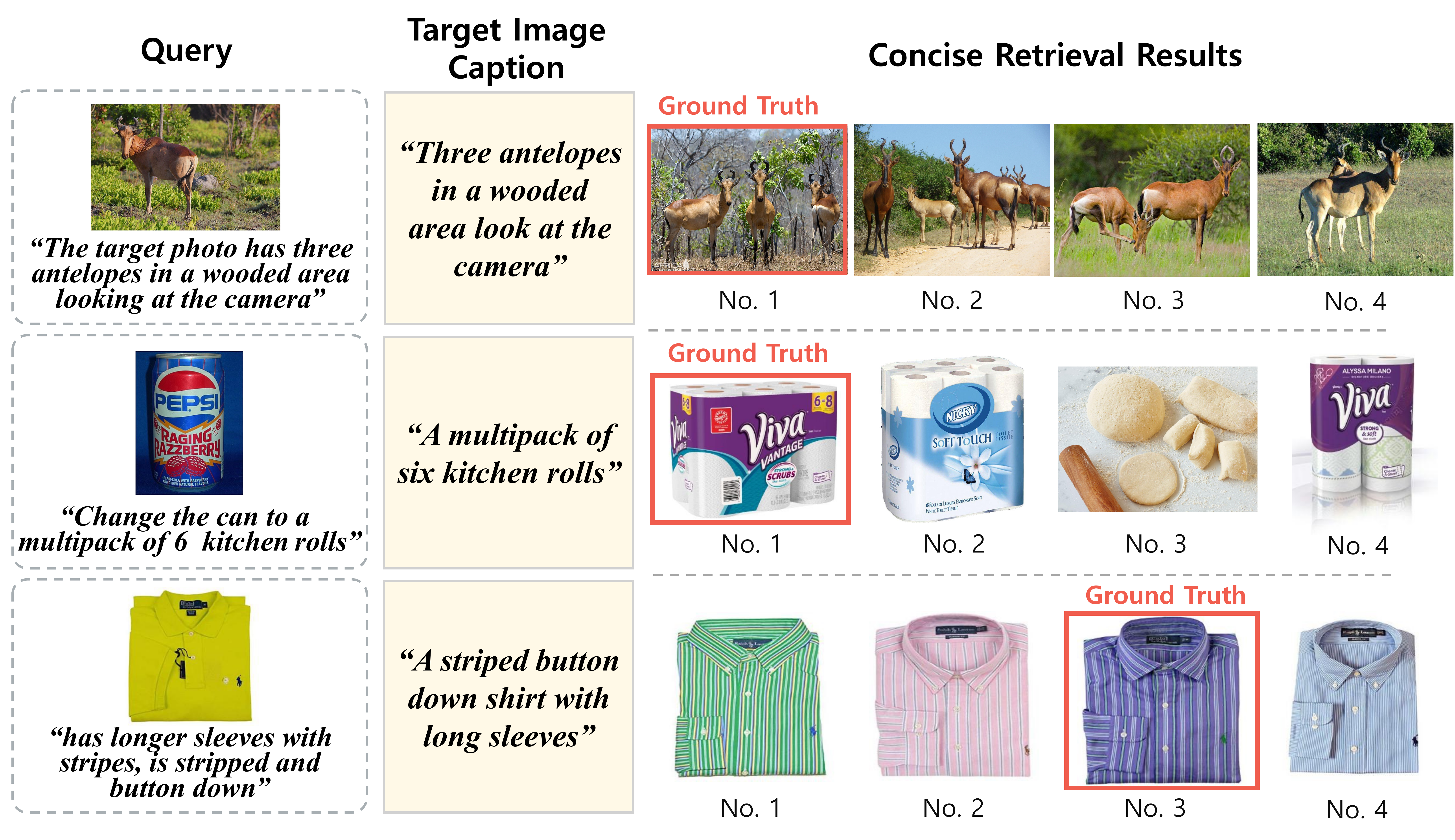}
    \caption{Examples of Concise Retrieval on the CIRR and FashionIQ validation benchmarks. Retrieval results are shown from left to right according to the ranking order, where the ground-truth target images are successfully retrieved at top ranks.}
    \Description{Examples of Concise Retrieval on the CIRR and FashionIQ validation benchmarks. Retrieval results are shown from left to right according to the ranking order, where the ground-truth target images are successfully retrieved at top ranks.}
    \label{fig:Qualitative_retrieval}
\end{figure}

\subsection{Qualitative Analysis}
\para{Concise Retrieval}
Figure~\ref{fig:Qualitative_retrieval} presents several qualitative examples of the proposed Concise Retrieval stage. Given a reference image and modification text, the LVLM can generate concise yet semantically accurate target image captions that effectively describe the desired retrieval target.
In the second example, the reference image contains a beverage can, while the modification text requests changing the object into a multipack of kitchen rolls. Instead of preserving irrelevant information from the reference image, the generated caption directly focuses on the target content: ``A multipack of six kitchen rolls.'' Notably, the original can is completely omitted from the caption. This demonstrates that the proposed concise caption generation strategy can recognize replacement-type modifications and suppress distracting reference semantics that are no longer relevant to the target images. The retrieval results are consistent with the generated caption, where the ground-truth image containing the kitchen rolls is ranked at the top position. Simultaneously, one retrieved candidate contains dough and rolling-pin-related visual details, suggesting that the phrase ``kitchen rolls'' may occasionally introduce semantic ambiguity during embedding-based retrieval.

The third example further demonstrates the robustness of the LVLM-based caption-generation process. The modification text contains the misspelled word ``stripped'' instead of ``striped.'' Nevertheless, the model correctly infers the intended meaning and generates the caption ``A striped button down shirt with long sleeves.'' This example highlights an important advantage of using LVLMs for caption generation: the model can interpret noisy or imperfect user instructions and convert them into semantically coherent, retrieval queries. Consequently, the retrieved candidates are all visually consistent striped shirts with different colors, and the ground-truth image is ranked among the top results.

Overall, these examples demonstrate that the proposed Concise Retrieval stage can effectively suppress irrelevant semantics, handle noisy instructions, and generate retrieval-oriented captions that improve candidate quality.

\begin{figure}[t]
    \centering
    \includegraphics[width=\columnwidth]{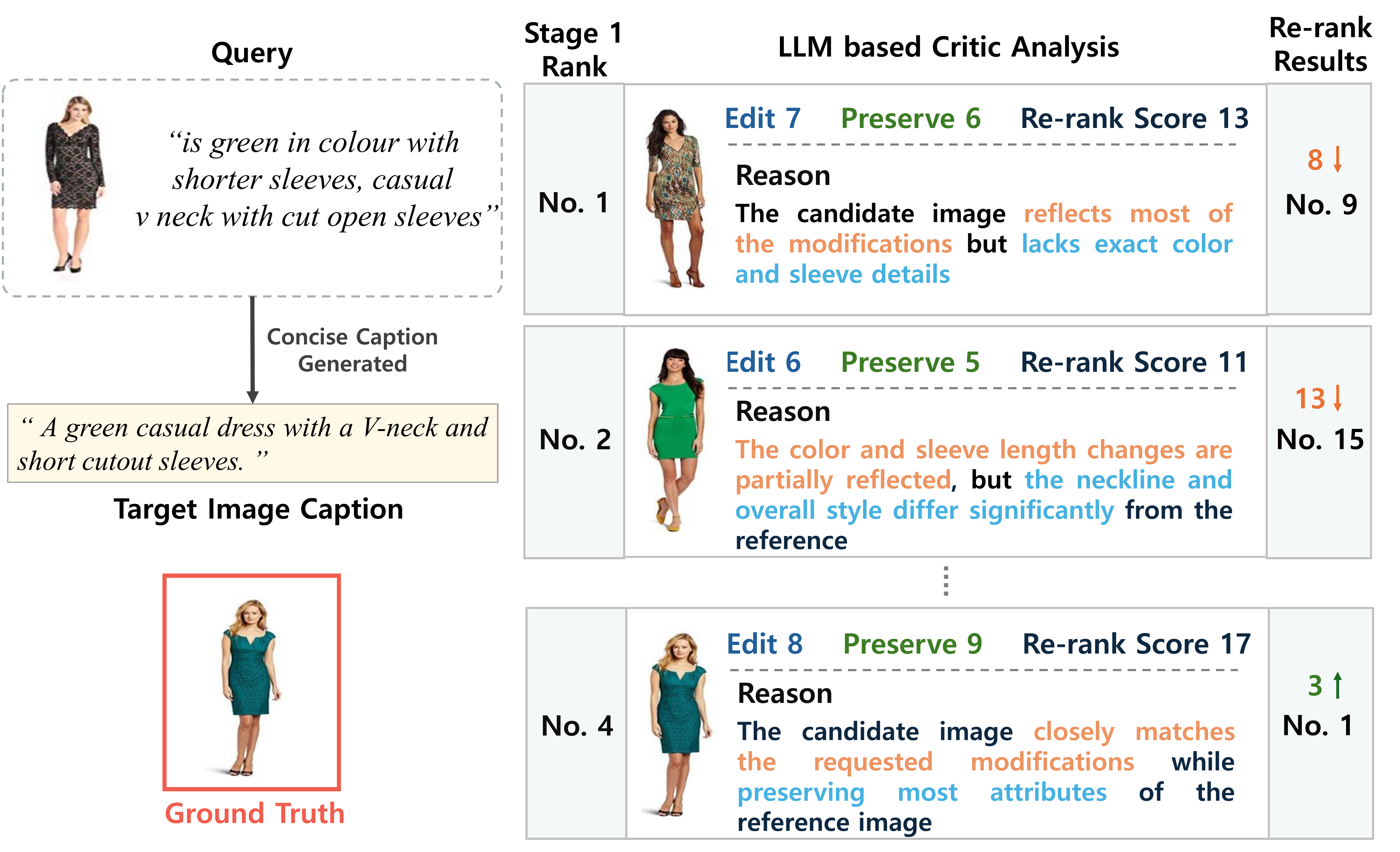}
    \caption{Example of the proposed critic-based re-ranking. The ground-truth target image, which is ranked low in the Stage-1 retrieval results, is successfully promoted to a higher rank after critic-based re-ranking.}
    \Description{
    This figure shows a qualitative example of critic-based re-ranking. 
    A reference image and modification text are used to generate the target caption. 
    Candidate images retrieved in Stage 1 are then evaluated by an LVLM critic using EDIT and PRESERVE scores. 
    The ground-truth candidate is promoted from rank 4 to rank 1 after re-ranking the list.
    }
    \label{fig:critic_rerank}
\end{figure}

\para{Critic Re-ranking}
Figure~\ref{fig:critic_rerank} illustrates how the proposed critic-based re-ranking further refines the preliminary retrieval results. In this example, the generated concise caption, ``A green casual dress with a V-neck and short cutout sleeves,'' correctly captures the intended target image semantics. However, the initial retrieval stage still produces imperfect rankings because of the incorporation of the image-image similarity term in the first-stage retrieval process. Specifically, because the reference image contains a patterned dress, the top-ranked candidate retrieved by the Concise Retrieval stage is also a patterned dress.

The critic successfully identifies this mismatch and assigns the initially top-ranked candidate a relatively unfavorable evaluation, explicitly pointing out that it ``lacks exact color and sleeve details.'' As a result, its ranking is significantly downgraded. Similarly, although the second-ranked candidate partially satisfies the color requirement, it fails to reflect the desired V-neck design. The critic recognizes that the neckline and overall style differ substantially from the intended target, leading to another downgrade in the final ranking.
In contrast, the ground-truth candidate originally ranked fourth achieves the highest critic score because it simultaneously satisfies the requested modifications while preserving most of the attributes of the reference image. The critic explicitly states that the candidate ``closely matches the requested modifications while preserving most attributes of the reference image,'' and the image is subsequently promoted to the top position.

These examples demonstrate that the proposed critic can complement embedding-based retrieval by explicitly evaluating both the modification consistency and reference preservation. While the first-stage retriever primarily captures coarse semantic similarity, the critic enables fine-grained visual reasoning over attributes such as color, sleeve design, neckline, and overall clothing style, resulting in more accurate final retrieval rankings.

\begin{table}[t]
\centering
\renewcommand{\arraystretch}{0.9}
\small 
\setlength{\tabcolsep}{2pt}
\caption{Ablation studies on the CIRCO test set using the ViT-L/14 backbone.}
\label{tab:ablation}
\resizebox{\columnwidth}{!}{
\begin{tabular}{l|cccc}
\toprule
\textbf{Methods} & \textbf{mAP@5} & \textbf{mAP@10} & \textbf{mAP@25} & \textbf{mAP@50} \\
\midrule

\multicolumn{5}{l}{\textbf{1. Critic Re-ranking Number}} \\
Top-5 
& 32.82 & 33.16 & 35.58 & 36.70 \\
Top-10 
& 36.82 & 35.47 & 37.86 & 38.98 \\
Top-20 (Full model) 
& \textbf{39.82} & \textbf{39.61} & \textbf{40.74} & \textbf{41.86} \\
\midrule

\multicolumn{5}{l}{\textbf{2. Critic Re-ranking LVLM}} \\
Qwen2.5-32B (Full model) 
& \textbf{39.82} & \textbf{39.61} & \textbf{40.74} & \textbf{41.86} \\
Qwen2.5-3B 
& 31.54 & 32.39 & 34.15 & 35.27 \\
Qwen2.5-7B 
& 35.87 & 36.19 & 37.62 & 38.74 \\
LLaVA-OneVision-7B 
& 21.84 & 23.08 & 25.96 & 27.07 \\
InternVL3-8B 
& 36.71 & 36.94 & 38.18 & 39.30 \\
Molmo-7B-D 
& 31.79 & 32.42 & 34.14 & 35.26 \\
\midrule

\multicolumn{5}{l}{\textbf{3. Critic Re-ranking $\lambda$}} \\
$\lambda = 0$ 
& 30.97 & 32.01 & 33.89 & 35.01 \\
$\lambda = 0.3$ 
& 34.81 & 35.57 & 36.96 & 38.07 \\
$\lambda = 0.5$ 
& 39.03 & 38.94 & 40.14 & 41.26 \\
$\lambda = 0.7$ (Full model) 
& \textbf{39.82} & \textbf{39.61} & \textbf{40.74} & \textbf{41.86} \\
$\lambda = 1$ 
& 37.67 & 37.72 & 38.98 & 40.09 \\
\bottomrule
\end{tabular}
}
\end{table}

\subsection{Ablation Study}
Extensive ablation studies are conducted on the CIRCO test set using the ViT-L/14 backbone to analyze the contribution of different components to the proposed framework. 

\para{Critic Re-ranking Number}
As shown in Table~\ref{tab:ablation}, reducing the re-ranking range noticeably degrades the performance. When only the Top-5 or Top-10 candidates are re-ranked, the retrieval performance consistently drops compared to the full Top-20 setting. This indicates that restricting the critic to a small candidate pool limits its ability to correct retrieval errors introduced during the first-stage retrieval.  It is worth noting that CIRCO adopts mAP as an evaluation metric, which evaluates the ranking quality across multiple relevant targets. Therefore, even when only the Top-5 candidates are updated, the performance changes can still propagate to higher evaluation ranges, such as mAP@50. Nevertheless, the influence is more significant on mAP@5 because the re-ranking operation directly modifies the highest-ranked samples.

\para{Critic Re-ranking LVLM}
Among all the evaluated models, Qwen2.5-32B achieves the best overall performance, whereas smaller variants, such as Qwen2.5-3B and Qwen2.5-7B, show clear performance degradation. Similar observations can also be found in models such as LLaVA-OneVision-7B and Molmo-7B-D, which exhibit significantly lower performances than the larger-scale Qwen models. This demonstrates that the semantic reasoning capability of the critic model is crucial for retrieval refinement. The performance gap can largely be attributed to the limited multimodal understanding ability of smaller LVLMs, which may fail to correctly understand the modified text and its relationship with the reference and target images, and may even fail to recognize textual content or objects within the images. 
Nevertheless, compared with the results in Table~\ref{tab:main_results}, even the smallest critic model, Qwen2.5-3B, consistently improves the Stage 1 Concise Retrieval results and surpasses all baseline methods across all metrics, demonstrating the effectiveness of the proposed critic-based re-ranking strategy, even with a lightweight LVLM.

\para{Critic Re-ranking $\lambda$}
The best performance is achieved at $\lambda = 0.7$, indicating that modification consistency contributes more to retrieval performance than reference preservation. This observation is reasonable because the CIR primarily focuses on retrieving images that correctly satisfy the modification instruction. Nevertheless, the performance drops when using only EDIT ($\lambda = 1$) or only PRESERVE ($\lambda = 0$), suggesting that both factors are necessary for a reliable retrieval ranking. Relying solely on EDIT may produce candidates that satisfy the modification text but deviate significantly from the reference image, while relying only on PRESERVE tends to favor visually similar images without adequately applying the desired modification. Therefore, balancing both objectives is important for achieving a robust retrieval performance.

\para{Caption Generation Strategies}
 All other retrieval components are removed in this experiment, and retrieval is performed under a pure text-image retrieval setting, where only the generated target image caption is used for retrieval. As shown in Table~\ref{tab:caption_compare}, existing methods such as CIReVL, OSrCIR, and AutoCIR are also included for comparison, as they similarly rely solely on caption-based similarity retrieval. Under the proposed concise prompting strategy, both GPT-5.1 and Qwen3 achieve better retrieval performance than previous methods. These findings suggest that concise caption generation can produce more discriminative retrieval semantics than verbose description-based prompting strategies. Compared with Qwen3, GPT-5.1 achieves better overall performance, which can likely be attributed to its stronger capability to follow concise caption generation instructions and understand both the image content and modification text.

\begin{table}[t]
\centering
\small
\renewcommand{\arraystretch}{0.9}
\setlength{\tabcolsep}{2pt}
\caption{Comparison of different caption generation strategies on the CIRCO test set under pure text-image retrieval.}
\label{tab:caption_compare}
\resizebox{\columnwidth}{!}{
\begin{tabular}{l|cccc}
\toprule
\textbf{Methods} & \textbf{mAP@5} & \textbf{mAP@10} & \textbf{mAP@25} & \textbf{mAP@50} \\
\midrule
Concise (GPT-5.1) & \textbf{25.28} & \textbf{26.13} & \textbf{28.49} & \textbf{29.53} \\
Concise (Qwen3) & \underline{24.28} & \underline{25.47} & \underline{27.54} & \underline{28.47} \\
CIReVL & 18.57 & 19.01 & 20.89 & 21.80 \\
OSrCIR & 23.87 & 25.33 & 27.84 & 28.97 \\
AutoCIR & 24.05 & 25.14 & 27.35 & 28.36 \\
\bottomrule
\end{tabular}
}
\end{table}

\begin{figure}[t]
    \centering
    \includegraphics[width=\columnwidth]{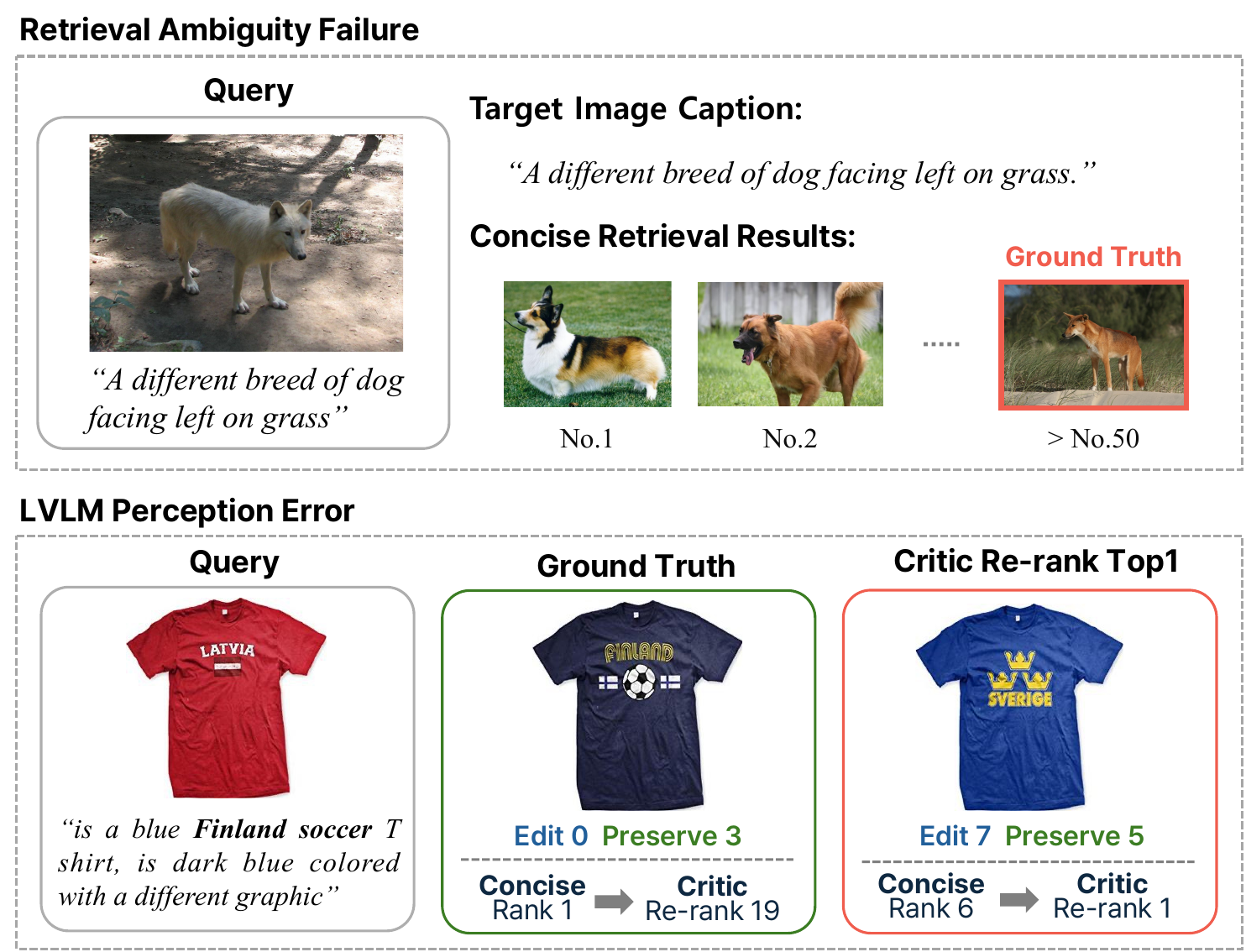}
    \caption{Failure cases of Concise Retrieval and Critic Re-ranking.}
    \Description{Failure cases of Concise Retrieval and Critic Re-ranking.}
    \label{fig:failure_retrieval}
\end{figure}

\subsection{Failure Case Analysis}

We further analyze the failure cases of our method. 
The failures are mainly attributed to two factors: retrieval ambiguity failure and LVLM perception errors.

\para{Retrieval Ambiguity Failure}
The first type of failure occurs during the concise retrieval stage, where the system fails to retrieve the ground-truth image within the Top-50 candidates. As illustrated in Figure~\ref{fig:failure_retrieval}, the system generates the target caption ``A different breed of dog facing left on grass'' from the reference image and modification text. Although the caption correctly reflects the intended edit, the corresponding ground-truth image is still ranked beyond the Top-50 retrieval results.

This failure mainly arises from the ambiguity of the modification instructions. The reference image contains a dog, while the modification text only specifies ``a different breed of dog facing left on grass'' without indicating any particular breed. Consequently, the generated target caption must remain semantically generic rather than introducing unsupported breed-specific information. Under such conditions, many candidate images depicting dogs facing left on grass become equally plausible matches, causing numerous visually valid candidates to receive similar retrieval scores. This case highlights how semantically underspecified queries can introduce retrieval ambiguity and reduce the reliability of ranking.

\para{LVLM Perception Error}
The second type of failure arises during the critic re-ranking stage. 
In this case, the concise retrieval stage ranks the ground-truth image first, whereas critic re-ranking substantially lowers its rank.
This failure is primarily caused by limitations in the perceptual capability of the LVLM critic, particularly when fine-grained visual understanding is required. As illustrated in Figure~\ref{fig:failure_retrieval}, the modification text specifies ``a blue Finland soccer T-shirt.'' Although the ground-truth image satisfies this requirement, the critic assigns an edit score of 0 and explains that ``The candidate image does not reflect the requested modifications and has minimal preservation of the reference's attributes.''
The generated explanation suggests that the critic fails to correctly recognize the textual and symbolic content of the clothing.
Such failures are likely caused by factors such as small font sizes, stylized typography, or low visual saliency of the text region. 

We further observe that this type of perception error becomes more severe when using lightweight LVLM critics. 
This observation indicates that the reliability of the re-ranking stage is strongly dependent on the visual reasoning capability of the critic model.

\section{Conclusion}

In this study, we propose LM-CIR, a training-free framework for ZS-CIR that combines Concise Retrieval and Critic Re-ranking. The proposed concise prompting strategy reduces semantic noise and produces more discriminative retrieval queries, while the critic re-rank stage further improves retrieval accuracy through LVLM-based compositional reasoning. 

One limitation of the current framework is that the critic re-ranking stage is applied only to the Top-$N$ results retrieved in the first stage. 
Future work could explore more efficient large-scale re-ranking strategies that expand the candidate pool while maintaining retrieval efficiency.

\begin{acks}
This work was supported by the Institute of Information \& Communications Technology Planning \& Evaluation (IITP) grant funded by the Korea government (MSIT) (RS-2026-25616928; RS-2024-00436936; RS-2024-00436934; RS-2024-00425354; RS-2024-00419073).
\end{acks}

\appendix
\section{Appendix}
\subsection{Stage 1: Concise Retrieval Prompts} 

\begin{figure}[h]
    \centering
    \includegraphics[width=0.98\columnwidth]{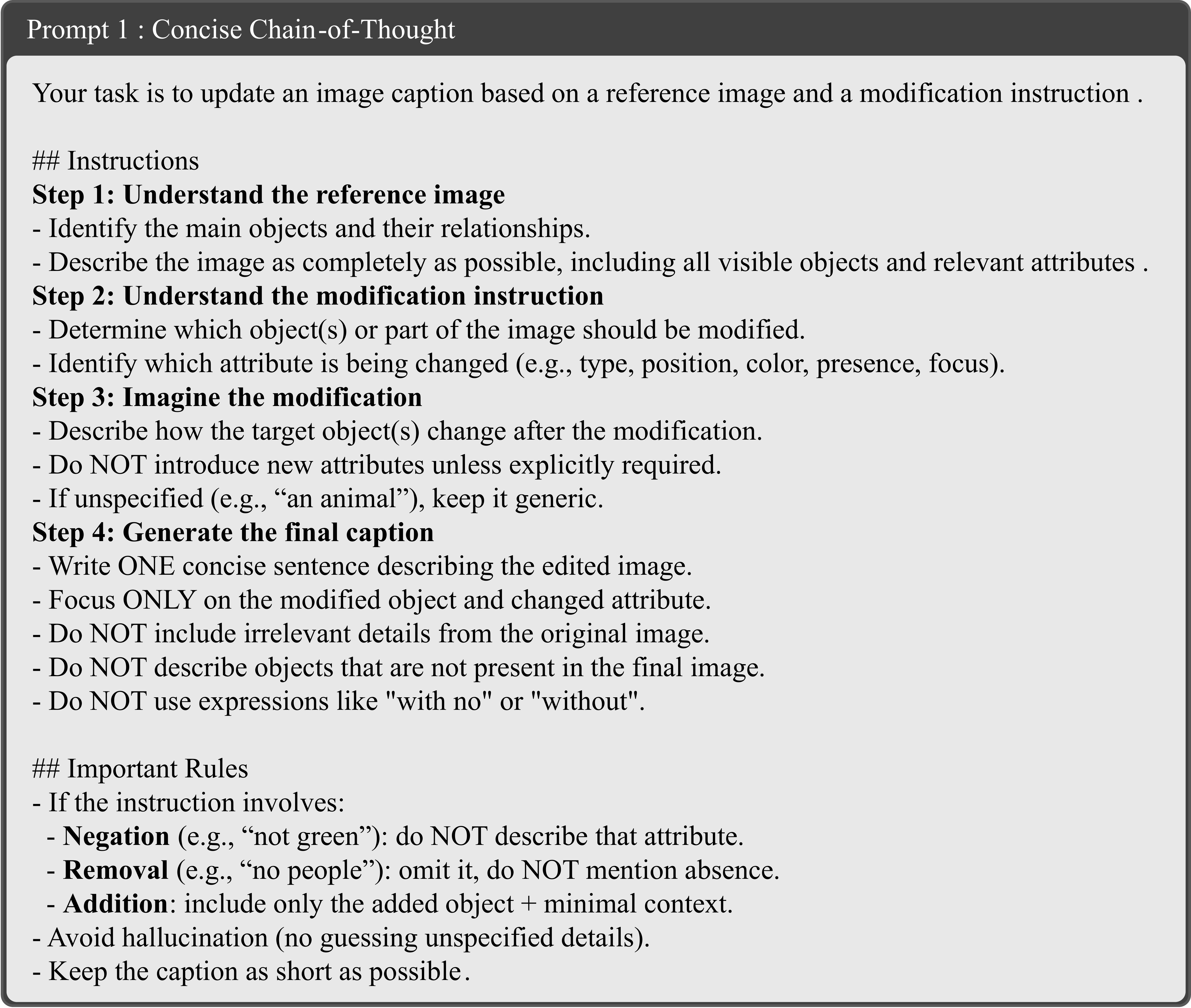}
    \caption{Concise Chain-of-Thought Prompting.}
    \Description{Concise Chain-of-Thought Prompting.}
    \label{fig:Prompt1}
\end{figure}

\begin{figure}[h]
    \centering
    \includegraphics[width=0.98\columnwidth]{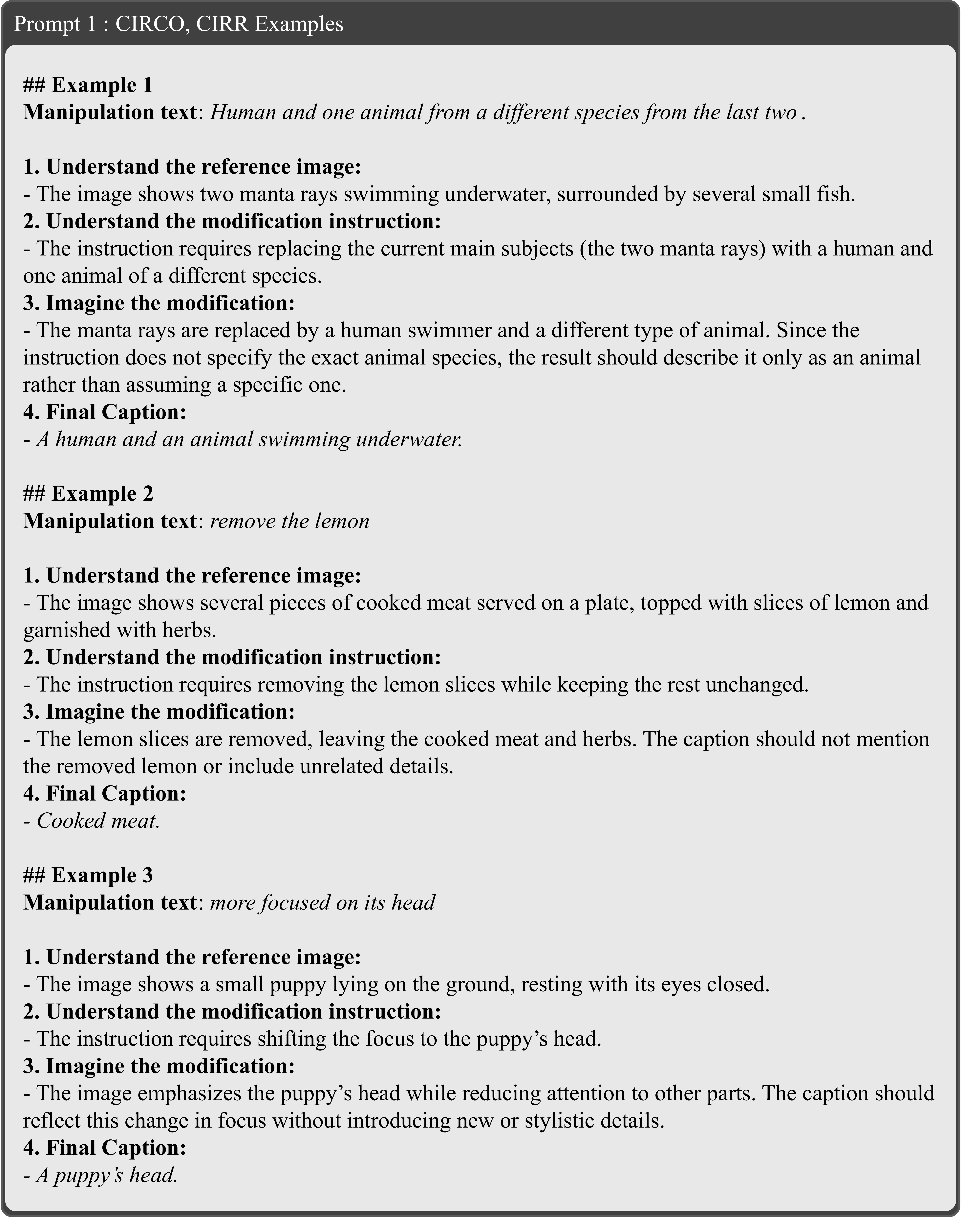}
    \caption{Prompt Examples for CIRCO and CIRR.}
    \Description{Prompt Examples for CIRCO and CIRR.}
    \label{fig:prompt_examples1}
\end{figure}

\begin{figure}[h]
    \centering
    \includegraphics[width=\columnwidth]{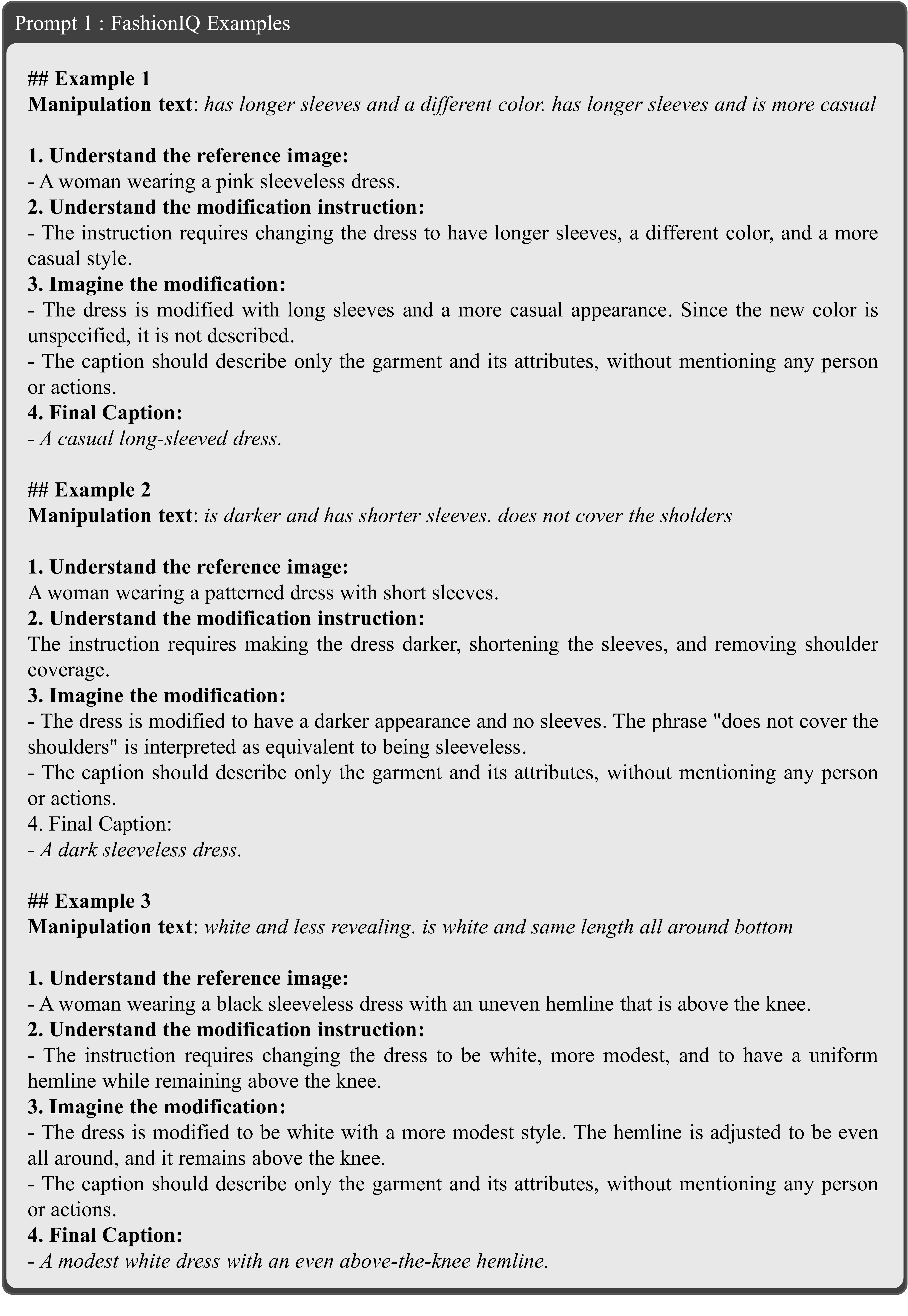}
    \caption{Prompt Examples for FashionIQ.}
    \Description{Prompt Examples for FashionIQ.}
    \label{fig:prompt_examples2}
\end{figure}

\subsection{Stage 2: Critic Re-ranking Prompt} 
\begin{figure}[h]
    \centering
    \includegraphics[width=\columnwidth]{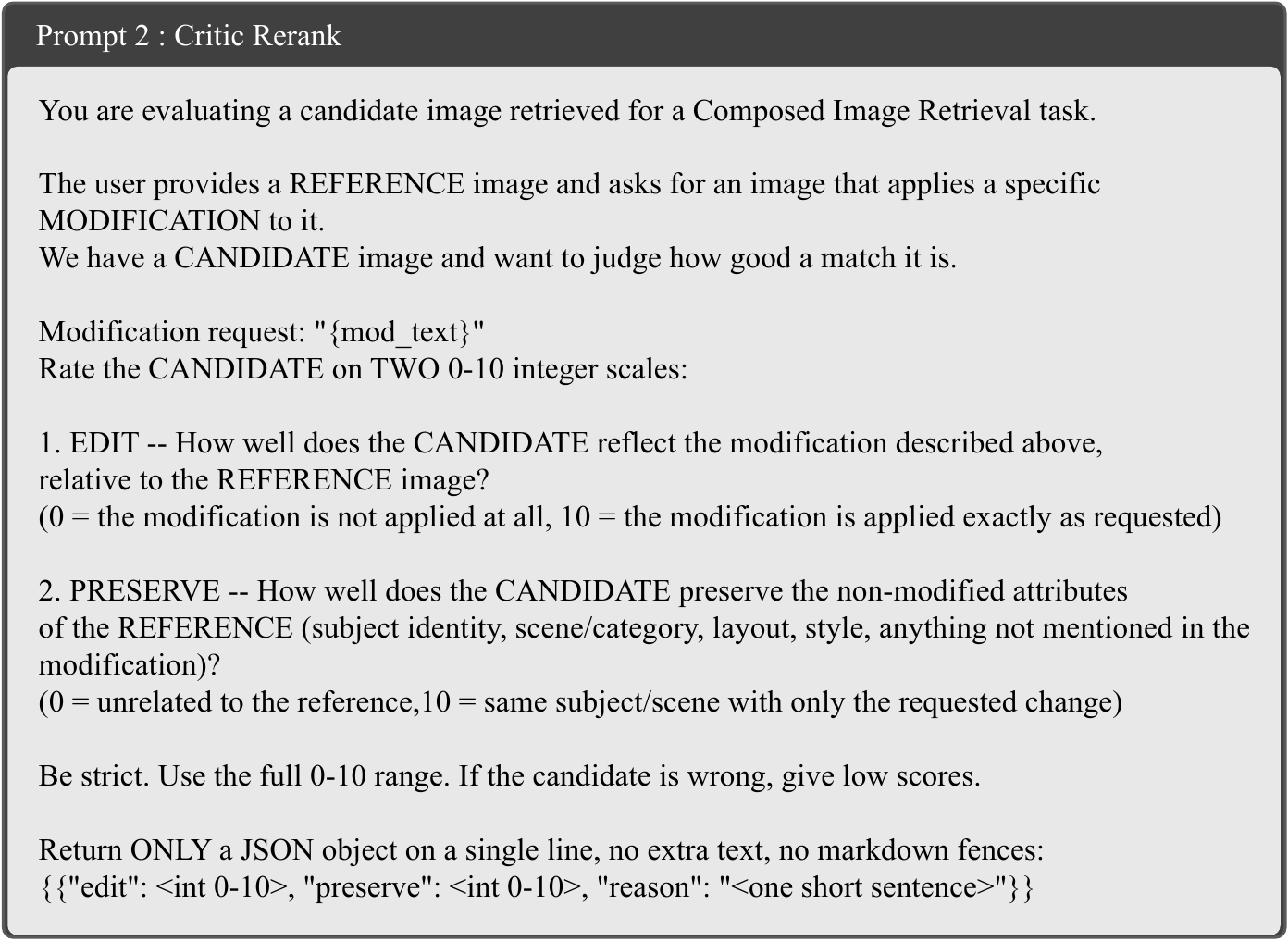}
    \caption{Critic Re-ranking Prompting.}
    \Description{Critic Re-rank Prompting.}
    \label{fig:Prompt2}
\end{figure}

\section*{GenAI Usage Disclosure}
Since this study is based on LVLMs, LVLMs were used in the experiments as described in the main text. GPT was additionally used for code checking and manuscript typo/grammar verification. All intellectual contributions, including code design and manuscript writing, were completed and verified by the authors under their direct supervision and responsibility.

\bibliographystyle{ACM-Reference-Format}
\balance
\bibliography{sample-base}

\end{document}